\documentclass{article}

\usepackage{enumitem}
\usepackage{amsmath,bm} 
\usepackage{subcaption} 
\usepackage{multirow}
\usepackage{arxiv}
\usepackage{amsmath}
\usepackage[ruled,vlined]{algorithm2e}
\usepackage[utf8]{inputenc} % allow utf-8 input
\usepackage[T1]{fontenc}    % use 8-bit T1 fonts
\usepackage{hyperref}       % hyperlinks
\usepackage{url}            % simple URL typesetting
\usepackage{booktabs}       % professional-quality tables
\usepackage{amsfonts}       % blackboard math symbols
\usepackage{nicefrac}       % compact symbols for 1/2, etc.
\usepackage{microtype}      % microtypography
\usepackage{lipsum}		% Can be removed after putting your text content
\usepackage{graphicx}

\usepackage{doi}
\usepackage[utf8]{inputenc}
\usepackage[numbers]{natbib} 
\DeclareMathOperator*{\argminA}{arg\,min} % Jan Hlavacek

\title{Semantic Image Fusion}
\author{{P.R. Hill} \\
	Visual Information Laboratory\\ University of Bristol, UK\\
	\texttt{paul.hill@bristol.ac.uk} 
	\And
	{D.R. Bull}\\
	Visual Information Laboratory\\ University of Bristol, UK\\
	\texttt{dave.bull@bristol.ac.uk}  \\
}

\renewcommand{\shorttitle}{\textit{arXiv} Template}

\begin{document}
\maketitle

\begin{abstract}
	Image fusion methods and metrics for their evaluation have conventionally used pixel based or low level features.  However, for many applications the aim of image fusion is to effectively combine the semantic content of the input images.  This paper proposes a novel system for the semantic combination of visual content using pre-trained CNN network architectures.  Our proposed semantic fusion is initiated through the fusion of the top layer feature map outputs (for each input image) through gradient updating of the fused image input (so called image optimisation).  Simple `choose maximum' and `local majority' filter based fusion rules are utilised for feature map fusion.  This provides a simple method to combine layer outputs and thus a unique framework to fuse single channel and colour images within a decomposition pre-trained for classification and therefore aligned with semantic fusion.  Furthermore, class activation mappings of each input image are used to combine semantic information at a higher level.  The developed methods are able to give equivalent low level fusion performance to state of the art methods while providing a unique architecture to combine semantic information from multiple images.
\end{abstract}

% keywords can be removed

\section{Introduction}
Image fusion is the combination of multiple images into a single image that aims to combine the most important visual information from all sources~\cite{hill2002image}.  Image fusion has been motivated by the need to improve visual representations, visualisation, scene understanding and situational awareness in multi-sensor and multi-camera applications such as remote sensing~\cite{GHASSEMIAN201675}, medicine~\cite{james2014medical} and surveillance~\cite{paramanandham2018multi}.

Image fusion has been driven by device and sensor limitations.  For example, not all important visual information can be captured by one type of sensor (e.g.\ IR, visible etc.) or within one single shooting setting (i.e.\ focus, angle etc.).  Furthermore, complementary imaging modalities co-exist within domains such as remote sensing and medicine that contain very different and important visual information.  The effective combination of all the visual information from all image sources is therefore the aim of image fusion.  Such a combined image is often  effective for subsequent tasks such as scene understanding and target recognition.

Image fusion has been a highly researched area over the last half century. During this time image fusion is performed at or decision-level, feature-level and pixel-level~\cite{li2017pixel}. Simple signal processing based pixel-level image fusion has given excellent results in preceding decades and continues to be used due to its high-efficiency and lack of a need for training data~\cite{li2017pixel}. Pixel-level fusion can be further classified into decomposition based methods or Sparse Representation (SR) based methods~\cite{li2017pixel}. In decomposition based techniques, the input images are decomposed into transform domains using methods such as complex wavelets (DT-CWT)~\cite{hill2002image}, the Discrete Wavelet Transform (DWT)~\cite{ben2005multiscale} and the contourlet transform~\cite{yang2010image}. Within the transform domain, coefficients are combined using suitably defined fusion rules (such as weighted-averaging \cite{jang2012contrast} and "choose maximum" \cite{hu2012multiscale}).   

More recently, state-of-the-art image fusion techniques have focused on the use of network based methods~\cite{javan2021review, hermessi2021multimodal,li2020nestfuse,li2018infrared,shao2018remote,tang2018pixel, amin2019ensemble,xu2020u2fusion}.  Due to the requirements of needing training data these methods are often domain focused with state-of-the-art results reported for IR/Visible fusion~\cite{li2020nestfuse, li2018infrared}, remote sensing (multispectral)~\cite{shao2018remote} and multi-focus areas~\cite{tang2018pixel, amin2019ensemble}.  Recent work has also been focused on universal image fusion methods that can effectively combine multiple sources within all of these domains~\cite{xu2020u2fusion}.  

There have only been a very small number of previously developed fusion methods that utilise semantic information for image fusion~\cite{fernandez2018remote, fan2019semantic}.  However, these methods do not use the joint classification and class activation maps to semantically fuse the input sources as proposed in our work.  These previous works also only use semantic information in a very limited way.

\subsection{Contributions}
The contributions and characteristics of our work are summarised as follows.
\begin{itemize}
    \item An unsupervised fusion technique is proposed that can combine the outputs of pre-trained low level feature maps using choose maximum and majority filter based fusion rules (using image optimisation).
    \item An unsupervised semantic fusion method that uses class activation maps.
    \item An unsupervised fusion method that combines both direct semantic fusion using class activation maps together with the combination of low-level network layer outputs (i.e.\ combining the above two approaches).
\end{itemize}

%We develop a new unsupervised network for image
%fusion by constraining the similarity between the
%fusion image and source images to overcome the
%universal stumbling blocks in most image fusion
%problems, i.e., the lack of universal ground truth and
%no-reference metric.
%• We release a new aligned infrared and visible image
%dataset, RoadScene, to provide a new option for image
%fusion benchmark evaluation. It is made available at
%https://github.com/hanna-xu/RoadScene.
%• We test the proposed method on six datasets for
%multi-modal, multi-exposure, and multi-focus image
%fusions. Qualitative and quantitative results validate
%the effectiveness and universality of U2Fusion.
\section{Feature Map Fusion using Image Optimisation}

Feature map based loss functions within pre-trained networks for image optimisation have been used extensively in Neural Style Transfer (NST) \cite{jing2019neural} and "Deep Dream" methods \cite{mordvintsev2015deepdream}.  This field was initiated by the seminal work in NST by Gatys et al.~\cite{gatys2015neural}.  Gatys' NST system applied previously developed texture synthesis methods \cite{gatys2015texture, portilla2000parametric} to the combination of two images (a "content" image and a "style" image) through an image optimisation method utilising a pre-trained Convolutional Neural Network (CNN): VGG19 \cite{simonyan2014very}.  Although these methods have provided amazing visual results they have yet to be utilised for general image processing applications such as image fusion and denoising.  NST using image optimised can be summarised through the generation of the output style transferred image ($I^*$) using the minimisation of a loss function: 

\begin{align} 
I^* &= \argminA_I \mathcal{L}_{total}(I| I_c, I_s)\label{equation1} \\
 &= \argminA_I \alpha \mathcal{L}_{c}(I| I_c) + \beta \mathcal{L}_{s}(I| I_s) 
\end{align}

\noindent where $I^*$ is the output image, $I_s$ is the "style" image and $I_c$ is the "content" image and $\alpha$ and $\beta$ are the loss weighting parameters.  The content and style losses ($\mathcal{L}_{c}, \mathcal{L}_{s}$) compare the content and style representations between the style and content images ($I_s$, $I_c$).  

As an initial step, each of these two images are decomposed into the layer outputs of a VGG19 CNN network~\cite{simonyan2014very}.  The two losses are calculated as functional comparisons between the layer outputs of the two images.  Image optimisation is then achieved through gradient updates through the network using the gradient on the image w.r.t. the total loss $\mathcal{L}_{total}$.  Since the original paper, many updates and optimisations to NST have been reported.  Li and Wand \cite{li2016combining} have proposed a Markov Random Field (MRF) based loss function that generates more plausible visual outputs.  Computational optimised NST methods include Johnson et al. \cite{johnson2016perceptual} and Ulyanov et al. \cite{ulyanov2016texture}.  These methods are similar to the Gatys' method in principle but are implemented using a single forward pass of a pre-trained network.  Multiple-Style-Per-Model NST methods have included Dumoulin et al. \cite{dumoulin2016learned}, Li et al. \cite{li2017diversified} and Zhang and Dana \cite{zhang2018multi}.  Finally GANs \cite{zhang2017style}, CycleGANs \cite{zhu2017unpaired} and image transformers \cite{deng2021stytr} have been recently used for NST.  Although there have been many advances in this field, the Gatys method is still considered to be the gold standard by most researchers  in terms of the quality of its results~\cite{jing2019neural}.  Therefore, although it is not computationally optimised or based on more complex transforms (GANs or Image Transformers), we have based our work on this type of image gradient update as it gives excellent results and is conceptually easy to understand and manipulate.  Computational optimisation of our developed methods can be implemented as future work. 

\subsection{Fusion Rules and Loss Functions for Image Optimisation based Image Fusion} 
The image optimisation methods described above have almost exclusively been used for Neural Style Transfer.  However, we propose such feature map image optimisation methods for image fusion.  For image fusion, a variety of fusion rules and loss functions have been tested over an exhaustive set of layer subsets of various pre-trained CNN networks.  It was found that taking losses across single layers gave the most effective results.  Furthermore, it is recognised that as the layers get further from the input images the semantic information increases.  This gives the possibility that fusion at such layers can combine more and more abstract semantic information.  However, such high level layers have a reduced resolution and therefore increased spatial support of each feature in the feature map.  It was found that fusing such layers generated unwanted artefacts such as banding.  Layers with the same resolution as the input images were found to give the most effective results (e.g.\ the first two layers of the VGG19 CNN).

Although very sophisticated fusion rules and loss functions have been considered, it was found that a simple choose maximum fusion rule combined with a $l_2$ loss function in most cases gave the best results.  The choice and utility of such a choose maximum fusion rule is motivated by its use within the wavelet transform domain \cite{hill2002image,ben2005multiscale} i.e.\ large magnitude coefficients (or feature map outputs) correlates with perceptually important content. 

Algorithm 1 illustrates the gradient update method used within the feature map based image fusion method.  This algorithm is based on the key concept of updating the input image $I^*$, input to a network $N$ through a gradient update with respect to a loss function such as (\ref{eq:lossFuse}) (as implemented within the style transfer methods and the deep dream method).  The loss function within the style transfer method is a weighted combination of the "content" and "style" losses defined through the comparison of the content and style image inputs.  The loss function within the DeepDream method~\cite{mordvintsev2015deepdream} is just based on the absolute magnitude of a chosen network output, layer or layer output.  Our fusion loss function is given in (\ref{eq:loss1}).  This is just the $L_2$ norm distance between the fused layer outputs (calculated just once outside the optimisation loops) and the iterated layer output feature map $F$.  The actual image fusion optimisation process is defined in (\ref{eq:min1}) being very similar to the NST equation (\ref{equation1}) (but with the input images being the images to be fused $I_0$ and $I_1$ rather than the style and content images: $I_s$ and $I_c$). 
\subsubsection{Loss Functions} 
The fused image $I^*$ is generated using image minimisation of the fusion loss function $\mathcal{L}_f$:
\begin{align}
I^* &= \argminA_I \mathcal{L}_f(I| I_0, I_1) \label{eq:min1} \\
\mathcal{L}_f(I| I_0, I_1) &= \sum_{l\in\{l_f\}} ||\left(\Psi \left((N^l(I_0),N^l(I_1)\right)-N^l(I)\right)||^2_2  \label{eq:loss1}
\end{align}
\noindent where $N^l(I)$ is the network feature map output at layer $l$ when the network is input image $I$. $l_f$ is the set of layers to calculate the loss over and $\Psi$ is the fusion rule (defined below).  Although an exhaustive combination of layers have been tested we have only utilised the top layer for our VGG19 CNN i.e.\ $l_f = \{'conv1\_1'\}$.
\subsubsection{Fusion Rules} 
Our initial fusion rule $\Psi_0$ for combining the layer outputs is the choose maximum rule:
\begin{align}
\Psi_0(F_0,F_1) = max\{F_x: x = 0...1\} \label{eq:chooseMax}
\end{align}
\noindent where $F_0$ and $F_1$ are the network outputs for the chosen layer (or set of layers) for input images 0 and 1 respectively. The maximum operator operates on a feature by feature basis in the selected output feature map.  The output is therefore a tensor the same size as the two input tensors where the tensor elements are selected according to the maximum absolute magnitude of the two corresponding tensor elements.

The choice between each of the spatial positions within the network layer outputs is effectively a binary decision mask.  This mask can be noisy and contain spatial inconsistencies.  We therefore use a localised majority filter as an alternative fusion rule $\Psi_1$.
\begin{align}
\Psi_1(F_0,F_1) = med_c\{|F_{0,\bm{i}+\bm{r},c}|>|F_{1,\bm{i}+\bm{r},c}|: \bm{r} \in W\} \label{eq:lossFuse}
\end{align}
\noindent $F_{0,\bm{i},c}$ is the feature map value at vector spatial position $\bm{i}$ and channel $c$ when when image $I_0$ is input into the network,  $\bm{r}$ is the spatial index vector of a sliding window ($3\times3$ in our case) and $med_c$ is the median value of the $3\times3$ boolean map on a channel by channel basis. This median filter applied on a boolean map therefore gives a spatial majority filter on a channel by channel basis for the considered feature map.

\begin{algorithm}[H]
\SetAlgoLined
\SetKwInOut{Input}{input}
\SetKwInOut{Output}{output}
\Input{Input Images: $I_0$, $I_1$}
\Output{Fused Image: $I^*$}
Initialise $I^*$: $I^* = mean(I_0,I_1)$\\
Fuse layer outputs: $G^l = \Psi \left((N^l(I_0),N^l(I_1)\right)$\\
%\KwInput{I_0,I_1}
 \For{$e$ Epochs}{
  Update ADAM learning rate: $\lambda$\;
  \For{$k$ Iterations}{
   Calculate layer feature loss: $\mathcal{L}_f(I^*| I_0, I_1) = \sum_{l\in\{l_f\}} ||\left(G^l-N^l(I^*)\right)||^2_2$ \\
   Update input image: $I^* = I^* + \lambda \nabla_{I^*} \left(\mathcal{L}_f(I^*| I_0, I_1)\right)$

   }{

  }
 }
 \caption{Layer-Based Fusion using Gradient Update: We k = 100, e = 100,
in our experiments.}
\end{algorithm}

\begin{figure*}
\centering
\begin{subfigure}[b]{.142\linewidth}
  \centering
  \includegraphics[width=\linewidth]{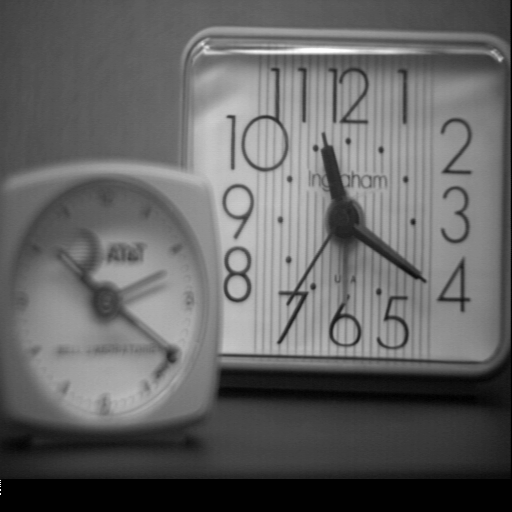}
\end{subfigure}%
\begin{subfigure}[b]{.142\linewidth}
  \centering
  \includegraphics[width=\linewidth]{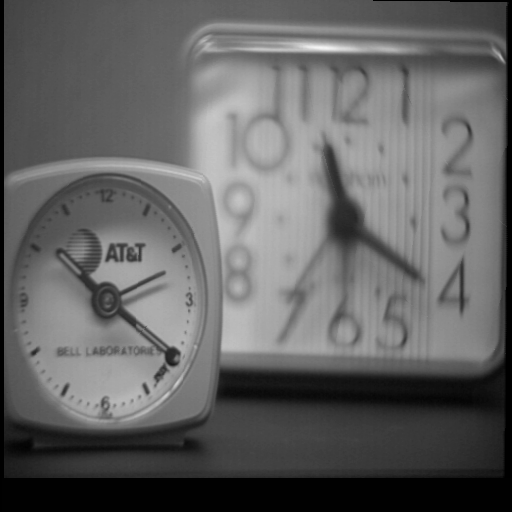}
\end{subfigure}%
\begin{subfigure}[b]{.142\linewidth}
  \centering
  \includegraphics[width=\linewidth]{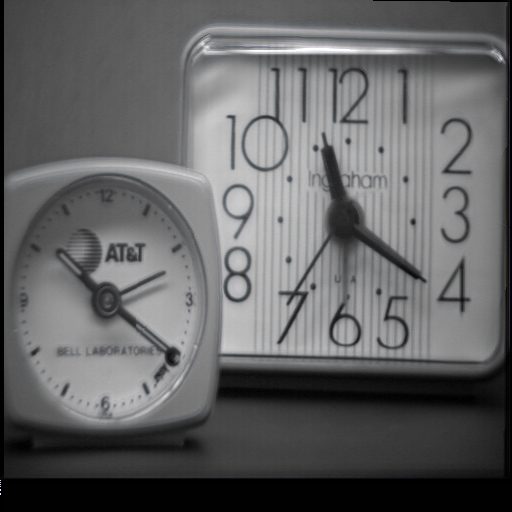}
\end{subfigure}%
\begin{subfigure}[b]{.142\linewidth}
  \centering
  \includegraphics[width=\linewidth]{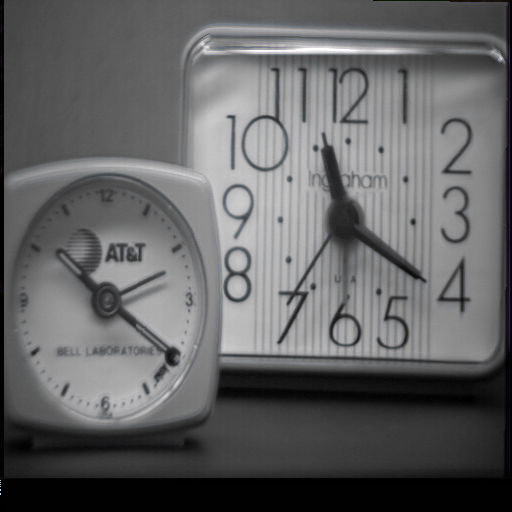}
\end{subfigure}%
\begin{subfigure}[b]{.142\linewidth}
  \centering
  \includegraphics[width=\linewidth]{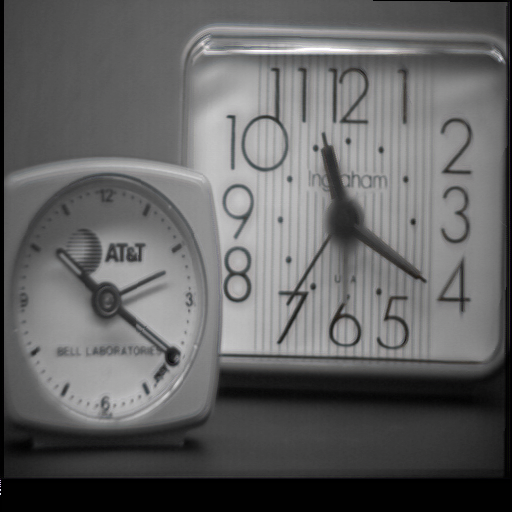}
\end{subfigure}%
\begin{subfigure}[b]{.142\linewidth}
  \centering
  \includegraphics[width=\linewidth]{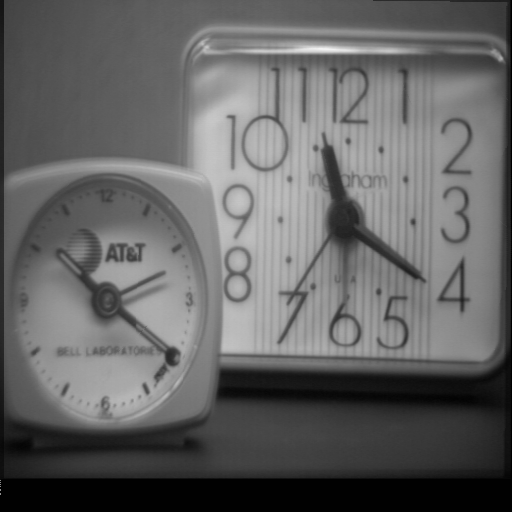}
\end{subfigure}%
\begin{subfigure}[b]{.142\linewidth}
  \centering
  \includegraphics[width=\linewidth]{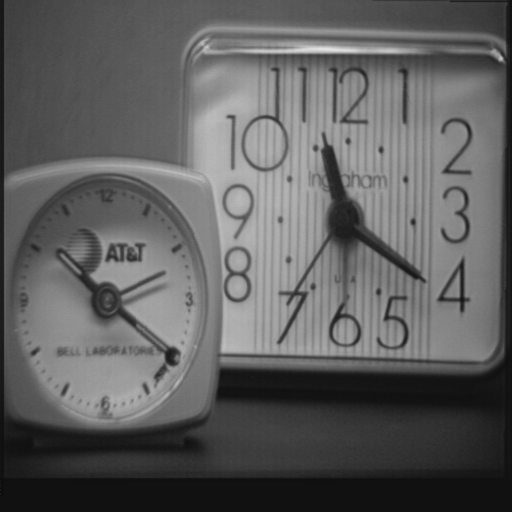}
\end{subfigure}
\\
\begin{subfigure}[b]{.142\linewidth}
  \centering
  \includegraphics[width=\linewidth]{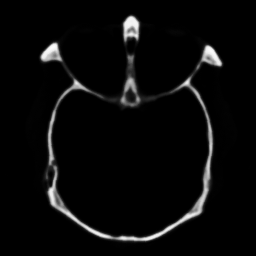}
\end{subfigure}%
\begin{subfigure}[b]{.142\linewidth}
  \centering
  \includegraphics[width=\linewidth]{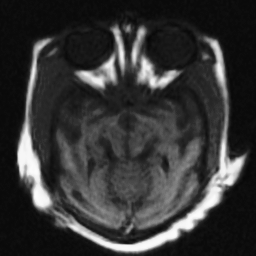}
\end{subfigure}%
\begin{subfigure}[b]{.142\linewidth}
  \centering
  \includegraphics[width=\linewidth]{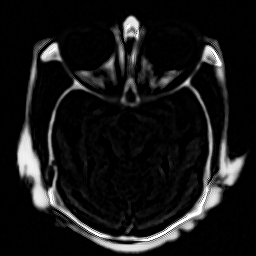}
\end{subfigure}%
\begin{subfigure}[b]{.142\linewidth}
  \centering
  \includegraphics[width=\linewidth]{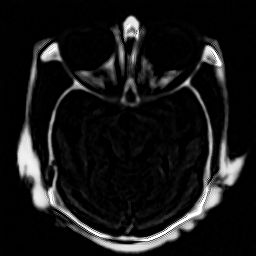}
\end{subfigure}%
\begin{subfigure}[b]{.142\linewidth}
  \centering
  \includegraphics[width=\linewidth]{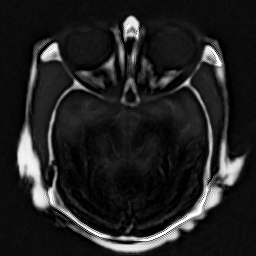}
\end{subfigure}%
\begin{subfigure}[b]{.142\linewidth}
  \centering
  \includegraphics[width=\linewidth]{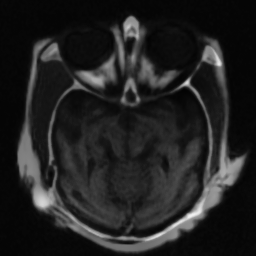}
\end{subfigure}%
\begin{subfigure}[b]{.142\linewidth}
  \centering
  \includegraphics[width=\linewidth]{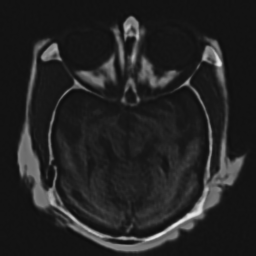}
\end{subfigure}
\\

\begin{subfigure}[b]{.142\linewidth}
  \centering
  \includegraphics[width=\linewidth, trim=0 0 2cm 0, clip]{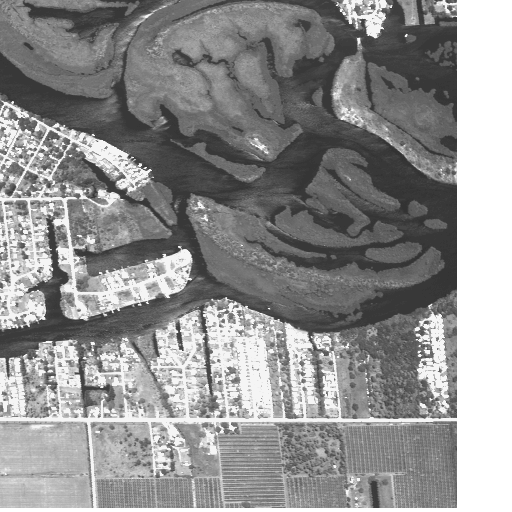}
\end{subfigure}%
\begin{subfigure}[b]{.142\linewidth}
  \centering
  \includegraphics[width=\linewidth, trim=0 0 2cm 0, clip]{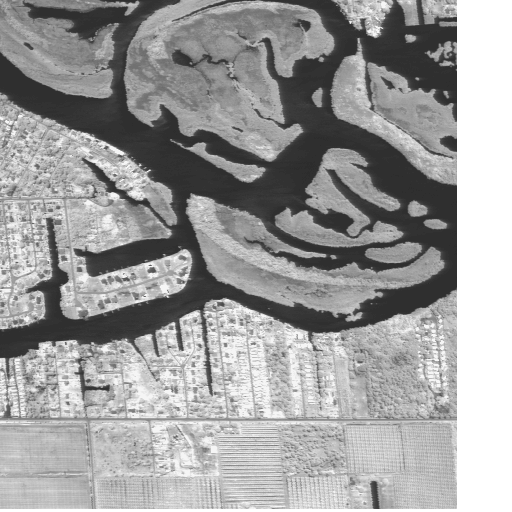}
\end{subfigure}%
\begin{subfigure}[b]{.142\linewidth}
  \centering
  \includegraphics[width=\linewidth,trim=0 0 2cm 0, clip]{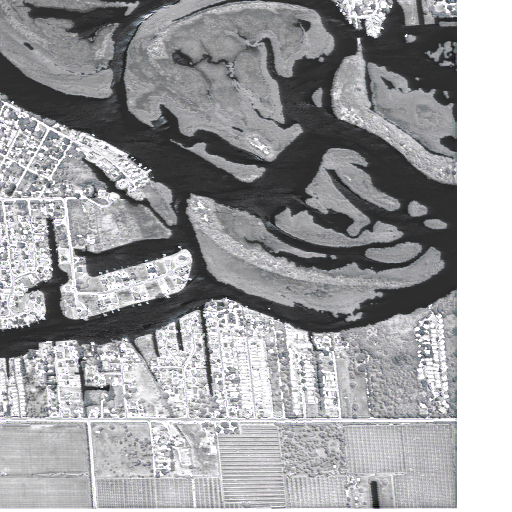}
\end{subfigure}%
\begin{subfigure}[b]{.142\linewidth}
  \centering
  \includegraphics[width=\linewidth, trim=0 0 2cm 0, clip]{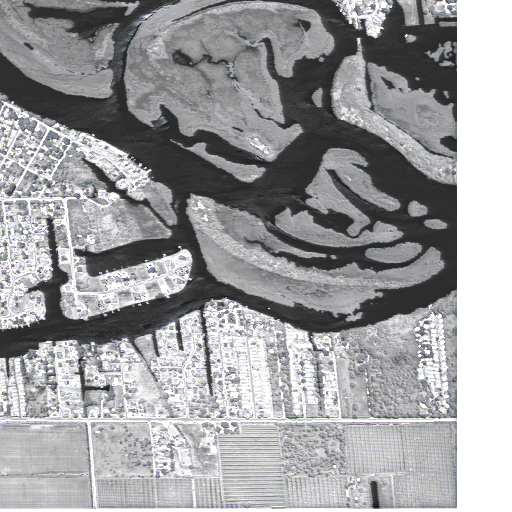}
\end{subfigure}%
\begin{subfigure}[b]{.142\linewidth}
  \centering
  \includegraphics[width=\linewidth, trim=0 0 2cm 0, clip]{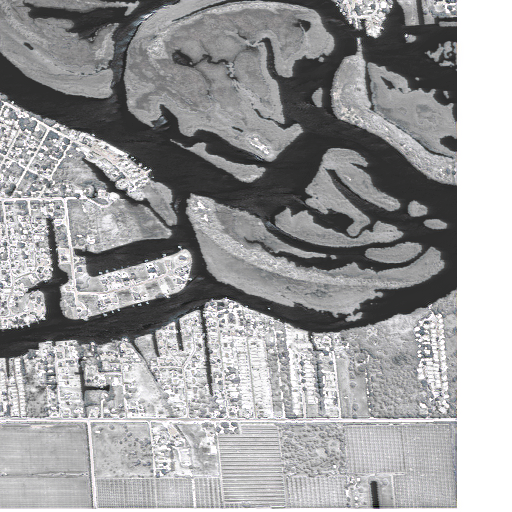}
\end{subfigure}%
\begin{subfigure}[b]{.142\linewidth}
  \centering
  \includegraphics[width=\linewidth, trim=0 0 2cm 0, clip]{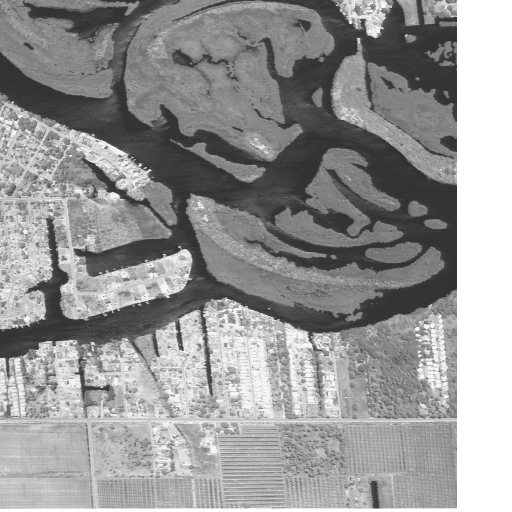}
\end{subfigure}%
\begin{subfigure}[b]{.142\linewidth}
  \centering
  \includegraphics[width=\linewidth, trim=0 0 2cm 0, clip]{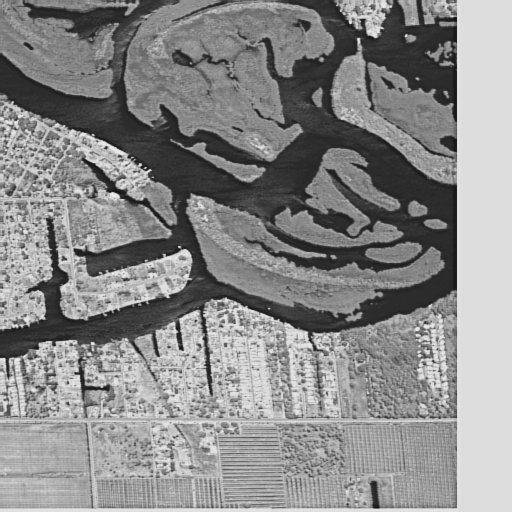}
\end{subfigure}\\

\begin{subfigure}[b]{.142\linewidth}
  \centering
  \includegraphics[width=\linewidth]{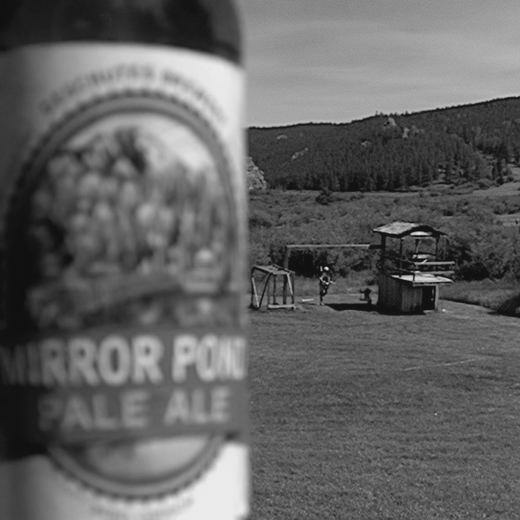}
\end{subfigure}%
\begin{subfigure}[b]{.142\linewidth}
  \centering
  \includegraphics[width=\linewidth]{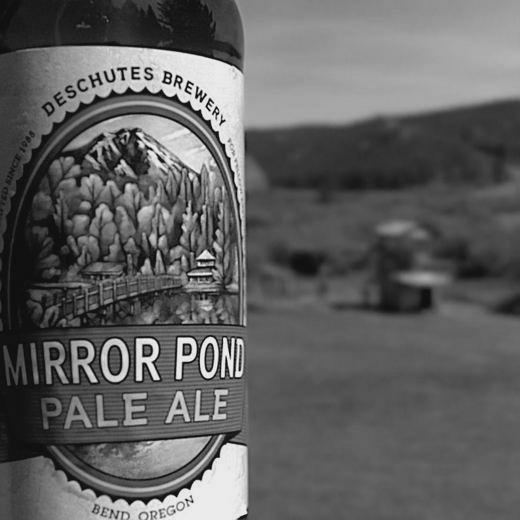}
\end{subfigure}%
\begin{subfigure}[b]{.142\linewidth}
  \centering
  \includegraphics[width=\linewidth]{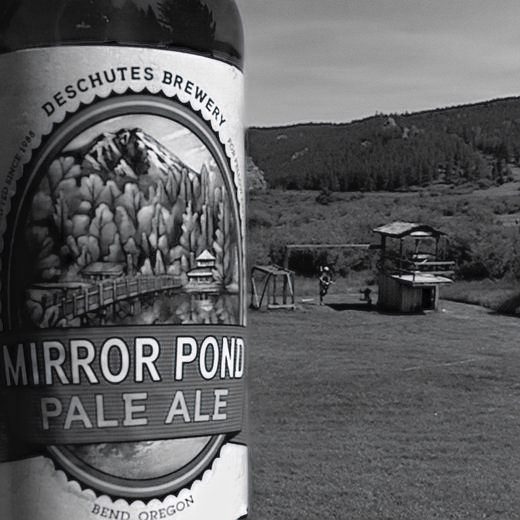}
\end{subfigure}%
\begin{subfigure}[b]{.142\linewidth}
  \centering
  \includegraphics[width=\linewidth]{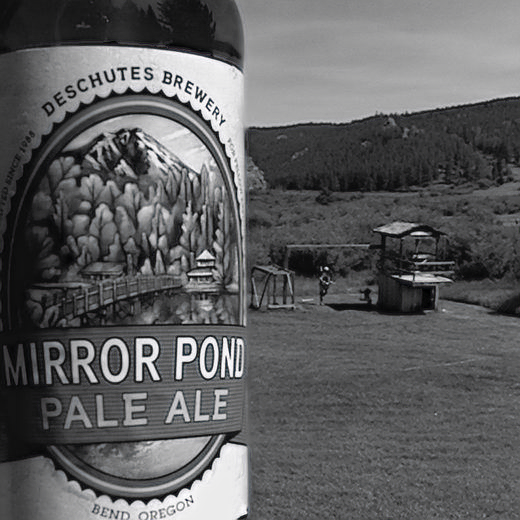}
\end{subfigure}%
\begin{subfigure}[b]{.142\linewidth}
  \centering
  \includegraphics[width=\linewidth]{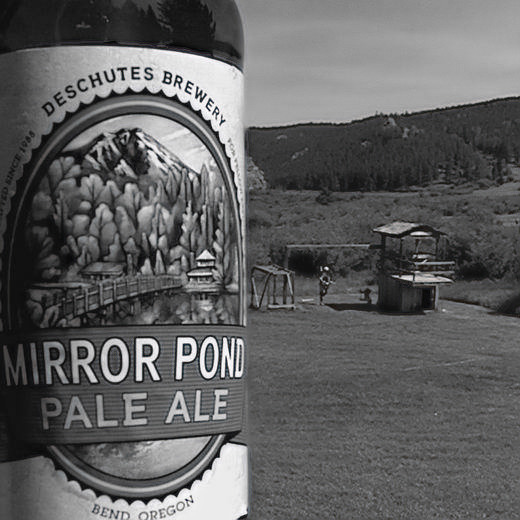}
\end{subfigure}%
\begin{subfigure}[b]{.142\linewidth}
  \centering
  \includegraphics[width=\linewidth]{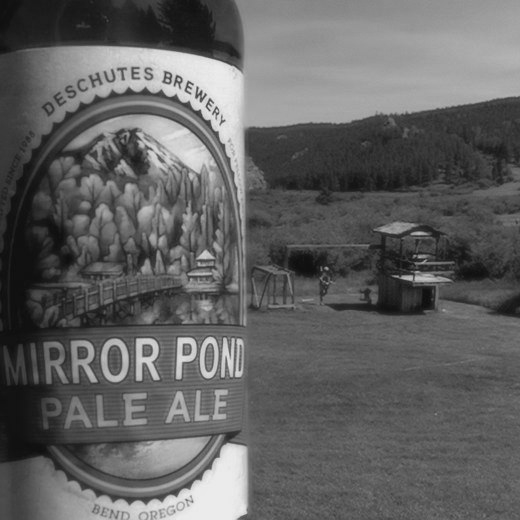}
\end{subfigure}%
\begin{subfigure}[b]{.142\linewidth}
  \centering
  \includegraphics[width=\linewidth]{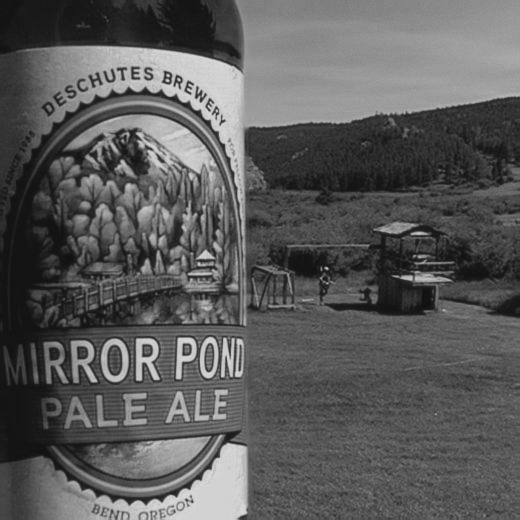}
\end{subfigure}
\\

\begin{subfigure}[b]{.142\linewidth}
  \centering
  \includegraphics[width=\linewidth]{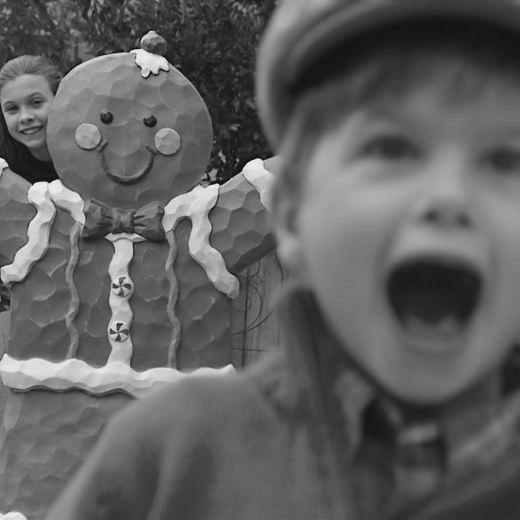}
\end{subfigure}%
\begin{subfigure}[b]{.142\linewidth}
  \centering
  \includegraphics[width=\linewidth]{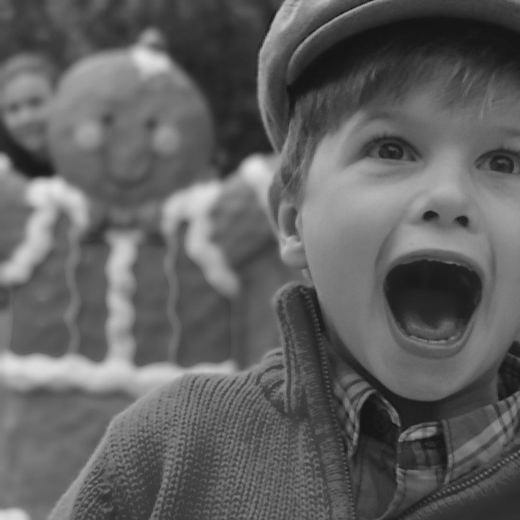}
\end{subfigure}%
\begin{subfigure}[b]{.142\linewidth}
  \centering
  \includegraphics[width=\linewidth]{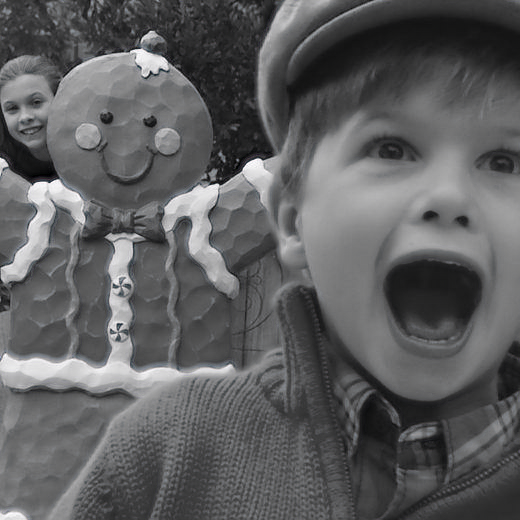}
\end{subfigure}%
\begin{subfigure}[b]{.142\linewidth}
  \centering
  \includegraphics[width=\linewidth]{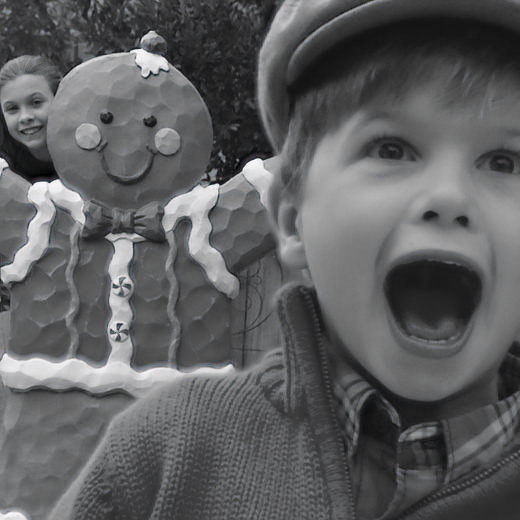}
\end{subfigure}%
\begin{subfigure}[b]{.142\linewidth}
  \centering
  \includegraphics[width=\linewidth]{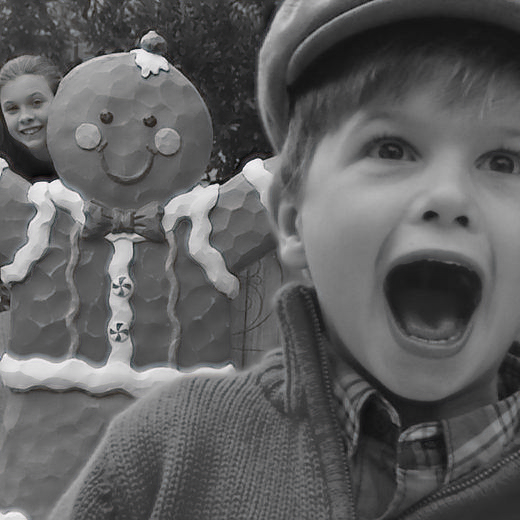}
\end{subfigure}%
\begin{subfigure}[b]{.142\linewidth}
  \centering
  \includegraphics[width=\linewidth]{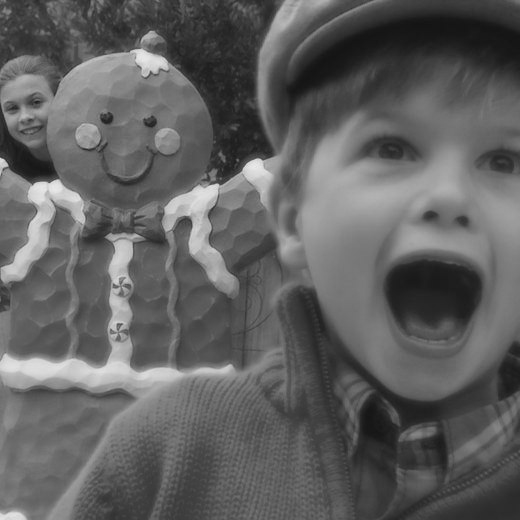}
\end{subfigure}%
\begin{subfigure}[b]{.142\linewidth}
  \centering
  \includegraphics[width=\linewidth]{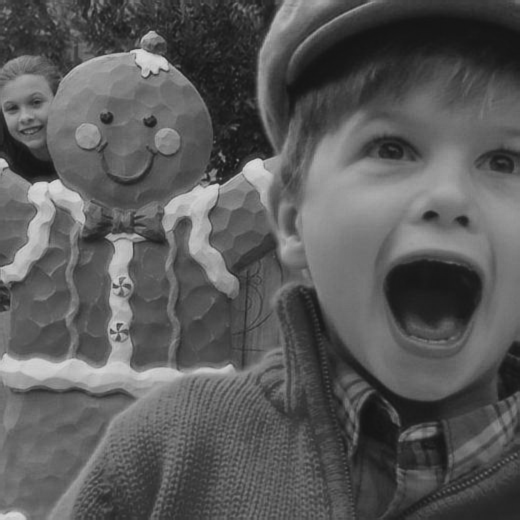}
\end{subfigure}\\

\begin{subfigure}[b]{.142\linewidth}
  \centering
  \includegraphics[width=\linewidth]{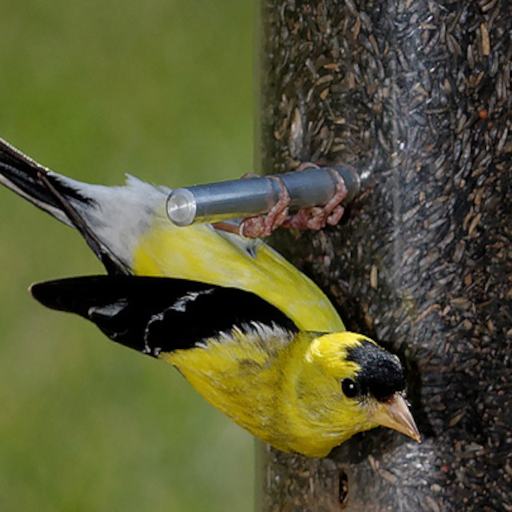}
\end{subfigure}%
\begin{subfigure}[b]{.142\linewidth}
  \centering
  \includegraphics[width=\linewidth]{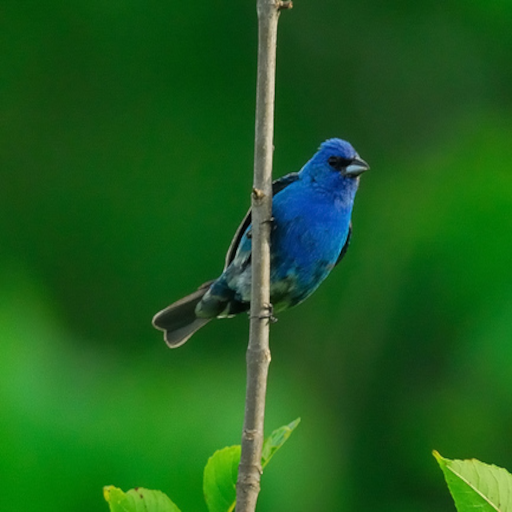}
\end{subfigure}%
\begin{subfigure}[b]{.142\linewidth}
  \centering
  \includegraphics[width=\linewidth]{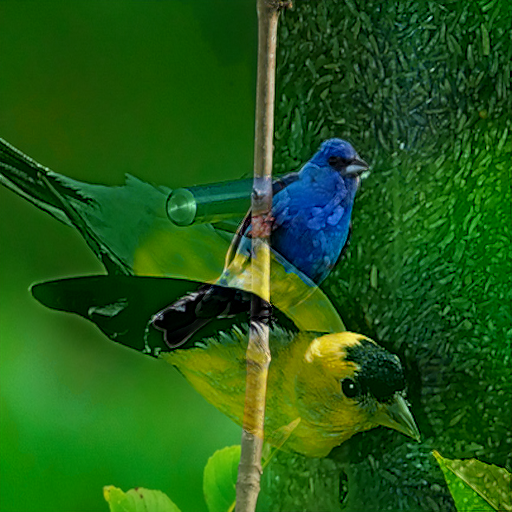}
\end{subfigure}%
\begin{subfigure}[b]{.142\linewidth}
  \centering
  \includegraphics[width=\linewidth]{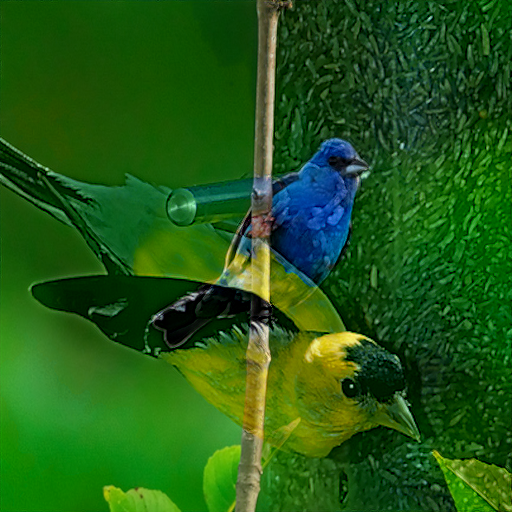}
\end{subfigure}%
\begin{subfigure}[b]{.142\linewidth}
  \centering
  \includegraphics[width=\linewidth]{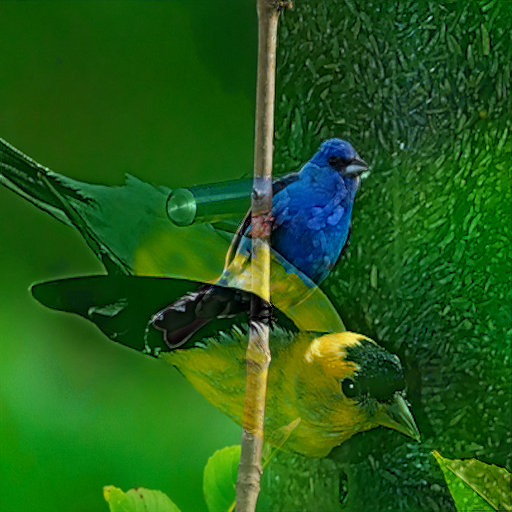}
\end{subfigure}%
\begin{subfigure}[b]{.142\linewidth}
  \centering
  \includegraphics[width=\linewidth]{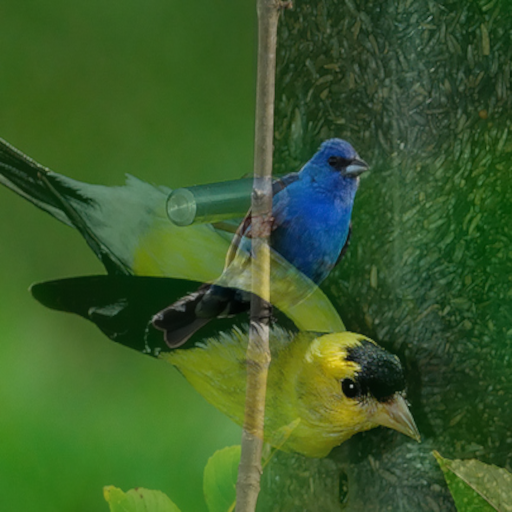}
\end{subfigure}%
\begin{subfigure}[b]{.142\linewidth}
  \centering
  \includegraphics[width=\linewidth]{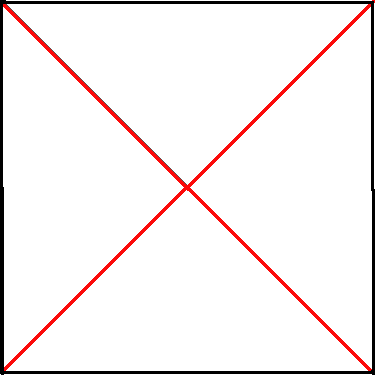}
\end{subfigure}\\

\begin{subfigure}[b]{.142\linewidth}
  \centering
  \includegraphics[width=\linewidth]{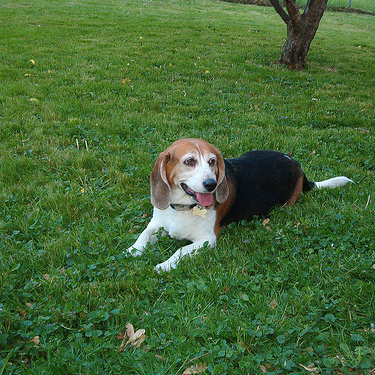}
\end{subfigure}%
\begin{subfigure}[b]{.142\linewidth}
  \centering
  \includegraphics[width=\linewidth]{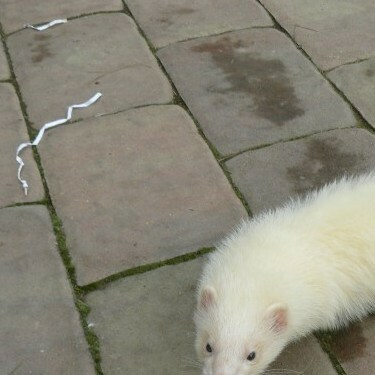}
\end{subfigure}%
\begin{subfigure}[b]{.142\linewidth}
  \centering
  \includegraphics[width=\linewidth]{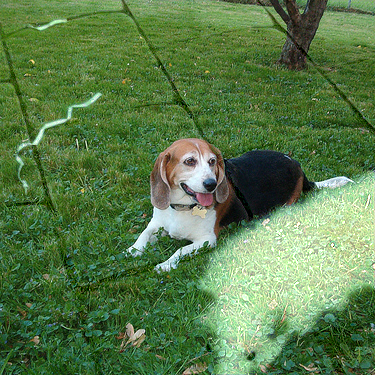}
\end{subfigure}%
\begin{subfigure}[b]{.142\linewidth}
  \centering
  \includegraphics[width=\linewidth]{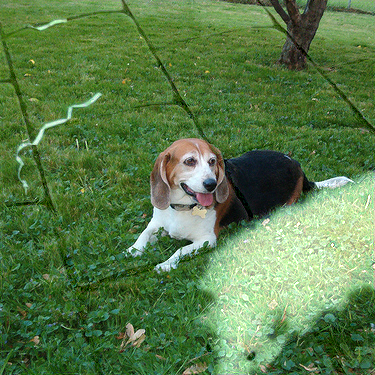}
\end{subfigure}%
\begin{subfigure}[b]{.142\linewidth}
  \centering
  \includegraphics[width=\linewidth]{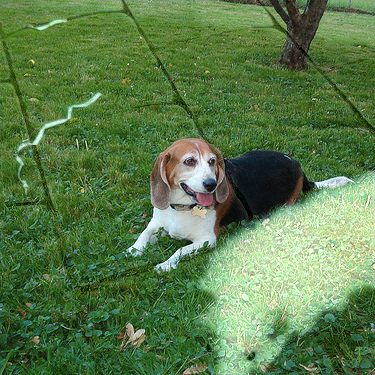}
\end{subfigure}%
\begin{subfigure}[b]{.142\linewidth}
  \centering
  \includegraphics[width=\linewidth]{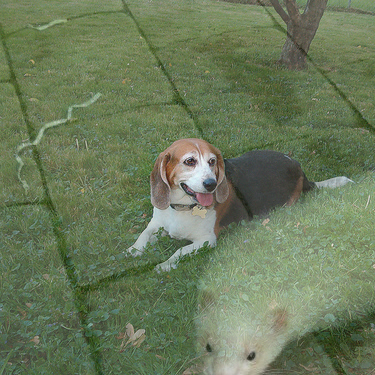}
\end{subfigure}%
\begin{subfigure}[b]{.142\linewidth}
  \centering
  \includegraphics[width=\linewidth]{./OutIms/blank.png}
\end{subfigure}\\
\begin{subfigure}[b]{.142\linewidth}
  \centering
  \includegraphics[width=\linewidth]{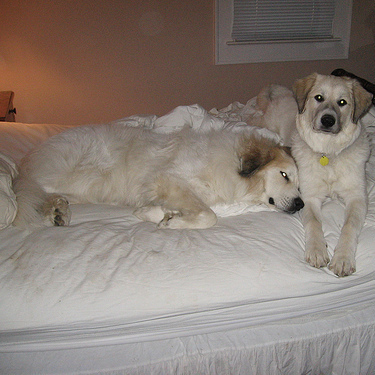}
  \caption*{Image 1}
\end{subfigure}%
\begin{subfigure}[b]{.142\linewidth}
  \centering
  \includegraphics[width=\linewidth]{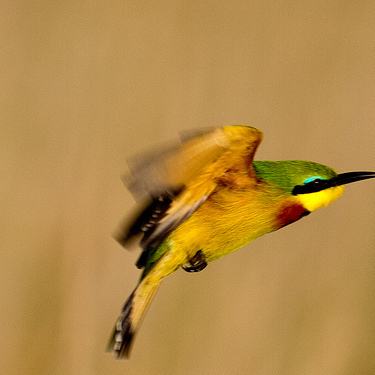}
  \caption*{Image 2}
\end{subfigure}%
\begin{subfigure}[b]{.142\linewidth}
  \centering
  \includegraphics[width=\linewidth]{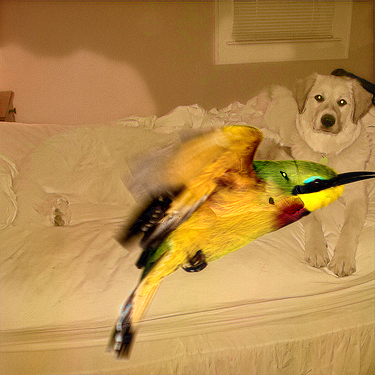}
\caption*{FM0}
\end{subfigure}%
\begin{subfigure}[b]{.142\linewidth}
  \centering
  \includegraphics[width=\linewidth]{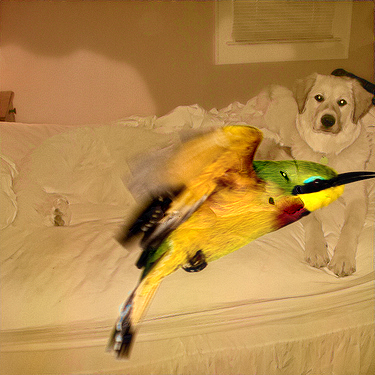}
\caption*{FM1}
\end{subfigure}%
\begin{subfigure}[b]{.142\linewidth}
  \centering
  \includegraphics[width=\linewidth]{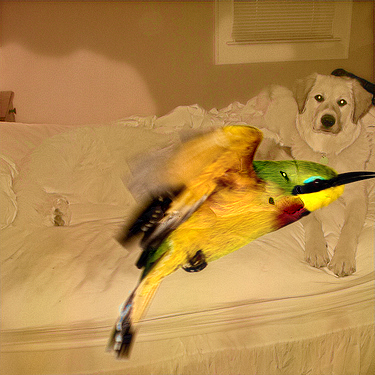}
\caption*{FM2}
\end{subfigure}%
\begin{subfigure}[b]{.142\linewidth}
  \centering
  \includegraphics[width=\linewidth]{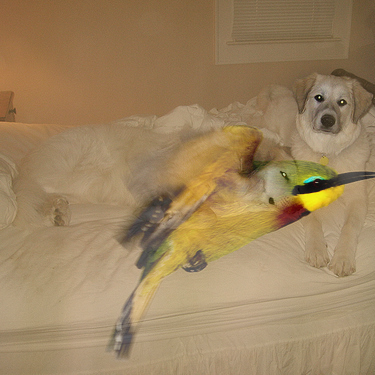}
\caption*{FM3}
\end{subfigure}%
\begin{subfigure}[b]{.142\linewidth}
  \centering
  \includegraphics[width=\linewidth]{./OutIms/blank.png}
\caption*{U2Fusion}
\end{subfigure}
\caption{Fused image results.  From top to bottom the fusion pairs are labelled
``Clock'', ``Head'', ``Remote'', ``Bottle'', ``Gingerbread'', ``Goldfinch/Indigo\_bunting'', ``Beagle/Ferret'', ``Great\_Pyrenees/Bee-Eater''.  U2Fusion results are not available for the last three rows as it is not possible to be used for colour images.}
\label{fig:fusion1}
\end{figure*}

\section{Semantic Fusion Using Class Activation Mappings (CAMs)}
The use of the more abstract layers within a CNN theoretically provides the ability to semantically fuse images using the image optimisation method described above.  However, due to these feature maps having very limited spatial resolution such fusion results in significant edge based artefacts.  We have therefore chosen to utilise Class Activation Mappings (CAMs) that give a spatial mapping of how relevant each pixel has been in generating a classification of a given class. 

\subsection{Class Activation Mappings}
 CAMs were first introduced by Zhou et al. \cite{zhou2016learning} and generate an importance spatial mapping according to a class $c$ through the sum of output weights of an input image for the last layer of a CNN.  This method was generalised to the use of backpropagated class gradients Grad-CAMs~\cite{selvaraju2017grad}.  These methods have been further extended using methods such as Eigen-CAM~\cite{muhammad2020eigen}, Grad-CAM using Vision-Transformers~\cite{aldahdooh2021reveal} etc.  However, for our work the mappings from Grad-CAM gave the best results (due to their spatial consistency).

Given the definitions of a Grad-CAM mapping by Selvaraju et al.~\cite{selvaraju2017grad} a spatial mapping (the same resolution as the input image) can be represented as $M_c^l(x,y)$ where $c$ is the class under consideration, $l$ is the analysed layer (usually the spatial layer closest to the actual classification layer).  We found that using the lowest resolution feature map, the mappings give the least spatial localisation.  Using higher resolution feature map layers gives better spatial localisation but less accurately reflects the abstract class activation's as the lower (more abstract) layers.  We therefore combine the lowest three spatial layers as follows:

\begin{align}
P_c(x,y) &= Norm\left(\prod_{l\in lset}M_c^l(x,y)\right)
\end{align}
\noindent where $Norm$ is the normalisation function:
\begin{align}
Norm(z) &= \frac{z - min(z)}{max(z)-min(z)}
\end{align}
\noindent As the $Norm$ function normalises the combination of the CAM mappings to the range [0,1] it can be considered as the probability $P_c$ of each pixel contributing to the classification of the top object recognised in each image (class $c_0$ for image $I_0$ and class $c_1$ for image $I_1$).  For our utilised VGG19 CNN, $lset$ is the set of the last three coarsest resolution layers i.e. $lset = \{'conv3\_4','conv4\_4', 'conv5\_4'\}$.

These probabilities are termed $P_{c_0,I_0}(x,y)$ and $P_{c_0,I_1}(x,y)$ for input images $I_0$ and $I_1$ at spatial positions $(x, y)$.  From these probabilities we need to generate a mixing ratio for the image fusion (also in the form of a probability termed $P_{M_0}(x,y)$: the probability that the output image should contain input image $I_0$).  $P_{M_0}(x,y)$ is generated as an exclusive combination of $P_{c_0,I_0}(x,y)$ and $P_{c_1,I_1}(x,y)$ i.e. 

\begin{itemize}[leftmargin=*]
    \item When $P_{c_0,I_0}(x,y)$ and $P_{c_1,I_1}(x,y)$ are approximately equal (for all values between 0 and 1), $P_{M_0}(x,y)$ should give an equal mix of each image in the output (i.e. $P_{M_0}(x,y)\approx 0.5$).
    \item When $P_{c_0,I_0}(x,y)$ is small (i.e.\ near 0) and $P_{c_1,I_1}(x,y)$ is large (i.e.\ near 1) then $P_{M_0}(x,y)$ should approximate 0).
    \item When $P_{c_1,I_1}(x,y)$ is small (i.e.\ near 0) and $P_{c_0,I_0}(x,y)$ is large (i.e.\ near 1) then $P_{M_0}(x,y)$ should approximate 1).
\end{itemize}

\begin{figure}[h]
\centering
\includegraphics[width=8cm]{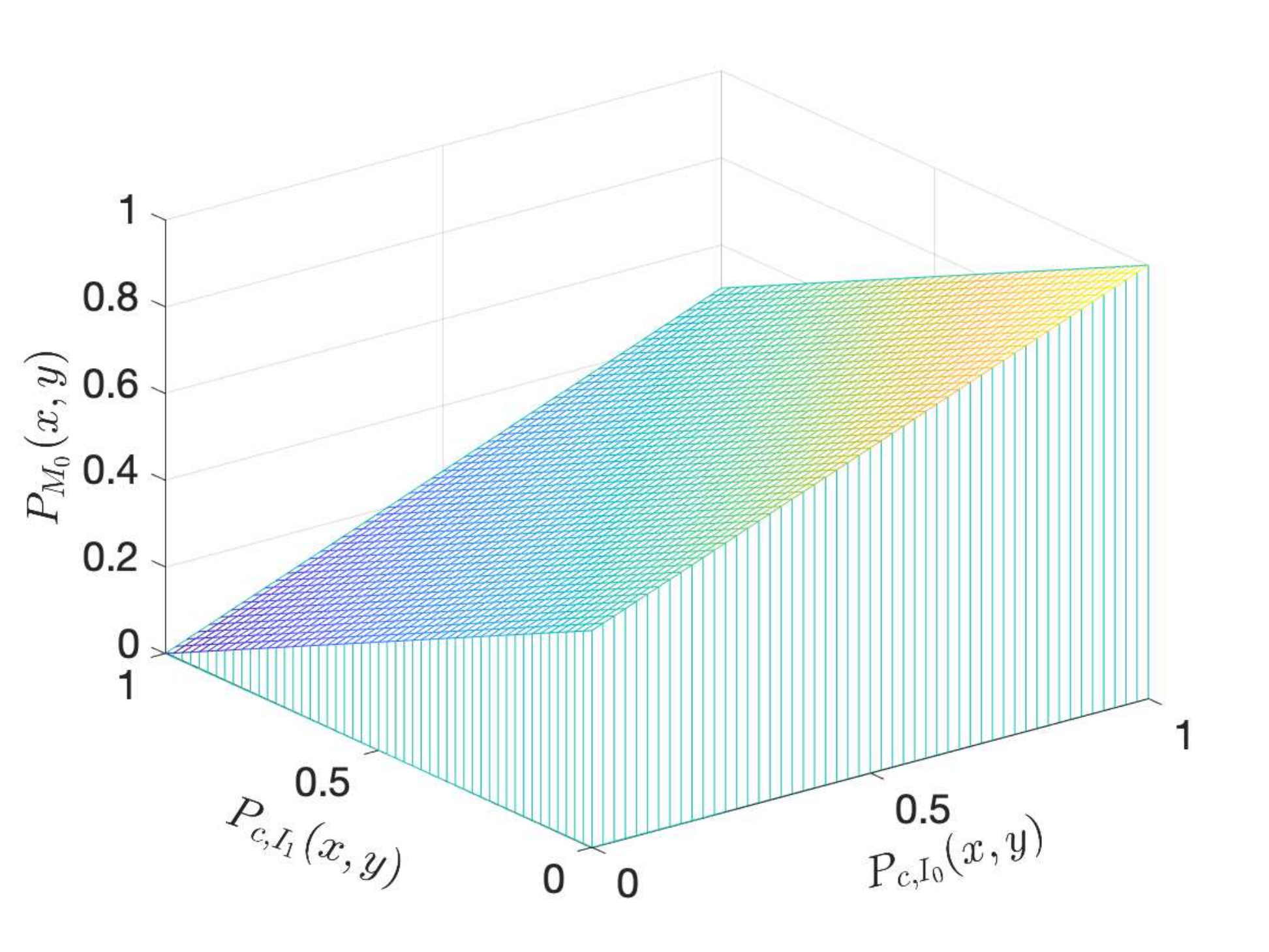}
\caption{Generating function (\ref{eq:mix}) to obtain mixing probability $P_{M_0}(x,y)$ from $P_{c_0,I_0}(x,y)$ and $P_{c_1,I_1}(x,y)$}
\label{fig:mix}
\end{figure}

This is achieved by defining $P_{M_0}(x,y)$ as (illustrated in figure \ref{fig:mix}):
\begin{align}
    P_{M_0}(x,y) = &0.5 (1+P_{c_0,I_0}(x,y)-P_{c_1,I_1}(x,y))\label{eq:mix}\\
    P_{M_1}(x,y) = & 1 - P_{M_0}(x,y)
\end{align}

Fusion is therefore achieved using $P_{M_0}(x,y)$ and $P_{M_1}(x,y)$ as the mixing ratios of the two input images.  The fused image $I^*$ is therefore defined as (dropping the $x,y$ indices).

\begin{align}
    I^* = &P_{M_0}I_0 + P_{M_1}I_1 
    \label{eq:mix1}
\end{align}

\section{Results}
Table 1 shows objective results for all the methods (defined below) for the single channel input images.  These have  been compared to state of the art methods U2Fusion \cite{xu2020u2fusion} and GMC \cite{anantrasirichai2020image}. The fusion metrics include: Wang ($Q_0$ \cite{wang2008performance}), Xydeas ($P_e$ \cite{2000Objective}) and Piella ($Q$ \cite{piella2003new}).  These metrics can only be effectively used for single channel images; colour images are therefore not included in this table.  This table shows that for the majority of the image pairs and metrics, the FM0 method (image optimisation based image fusion) gives the best performance.  The GMC method gives the best results for all metrics for the head image pair.  This is unsurprising as GMC was specifically designed to be used in this domain \cite{anantrasirichai2020image}.

\subsection{Method Definitions}
\begin{description}
    \item[FM0,] Fusion Method 0: This method utilises image optimisation using (\ref{eq:min1}) and (\ref{eq:loss1}) and fusion rule (choose maximum) $\Psi_0$ defined in (\ref{eq:chooseMax}).
    \item[FM1,] Fusion Method 1: This method utilises image optimisation using (\ref{eq:min1}) and (\ref{eq:loss1}) and fusion rule (majority filter) $\Psi_1$ defined in (\ref{eq:lossFuse}).
    \item[FM2,] Fusion Method 2: This combines method FM0 and FM3 (i.e. feature maps $F_0$ and $F_1$ are multiplied by CAM probabilities $P_{c_0,I_0}$ and $P_{c_1,I_1}$ respectively).
    \item[FM3,] Fusion Method 3: Fusion using CAMs utilising (7),(8),(9),(10) and (11).
\end{description}

\subsection{Classification for Semantic Image Fusion}
In order to generate the Class Activation Mappings (CAMs) the chosen network (VGG19) was used to get the top classification class $c$ for each of the input images i.e.\ class $c_0$ is the top classification class for input image $I_0$ and $c_1$ is the top classification class for input image $I_1$.  These classifications don't always make sense for all the considered applications and domains (e.g.\ the "head" and "remote"  images in Figure 1).  However, table 2 shows the top classification classes for the image pairs shown in figure 3.  Figure 3 also shows the CAM mappings for the top classes for each of the image pairs.  This figure shows how the different semantic objects are highlighted by the class activation mappings and how this is utilised to combine and semantically fuse the input images.  This figure shows how the utilisation of CAM mappings can distinguish between spatial regions that are in focus in multi-focus fusion pairs (see the clock CAMs in the top row).

\begin{table}[!htb]
\begin{center}
\small
\begin{tabular}{ccccc}
\hline
Dataset & Method & $Q_0$ & $P_e$ & $Q$\\ \hline

\multirow{4}{*}{Clock}& FM0  & 0.8294&    \textbf{0.6308}&    \textbf{0.9580}\\
& FM1  &    0.8293  &  0.6262  &  0.9572\\
& FM2  &    \textbf{0.8296}  &  0.6199  &  0.9544\\
& FM3  &    0.8293  &  0.5881  &  0.9495\\
& U2Fusion  &    0.8267  &  0.5409  &  0.8970\\
& GMC  &    0.8286  &  0.5208   & 0.9325\\ \hline             
\multirow{4}{*}{Head}& FM0  & 0.8035  &   0.3425  &   0.3045\\
& FM1  &    0.8034  &   0.3411  &   0.3049\\
& FM2  &    0.8045 &    0.5539 &    0.6614\\
& FM3  &    0.8066 &    0.4753  &   0.7117\\
& U2Fusion  &    0.8072  &   0.2732  &   0.7056\\
& GMC  &    \textbf{0.8216} &    \textbf{0.6660} &    \textbf{0.8002}\\ \hline
\multirow{4}{*}{Remote}& FM0  & 0.8100  &   \textbf{0.6490}  &   \textbf{0.8640}\\
& FM1  &    0.8103  &   0.6394  &   0.8604\\
& FM2  &    0.8106  &   0.6364  &   0.8597\\
& FM3  &    0.8121  &   0.5827  &   0.8544\\
& U2Fusion  &    0.8097  &   0.5458  &   0.8487\\
& GMC  &    \textbf{0.8156}  &   0.6210  &   0.7708\\ \hline
\multirow{4}{*}{Bottle}& FM0  & 0.8196  &   \textbf{0.7112}  &   \textbf{0.9491}\\
& FM1  &    0.8191  &   0.6993  &   0.9470\\
& FM2  &    \textbf{0.8199}  &   0.6960  &   0.9476\\
& FM3  &    0.8164  &   0.4688  &   0.8819\\
& U2Fusion  &    0.8168  &   0.6111 &    0.9286\\
& GMC  &    0.8166 &    0.4830  &   0.8725\\ \hline
\multirow{4}{*}{Gingerbread}& FM0  & 0.8217  &   \textbf{0.6681} &   \textbf{0.9522}\\
& FM1  &    0.8215  &   0.6633  &   0.9517\\
& FM2  &    \textbf{0.8230} &    0.6497  &   0.9489\\
& FM3  &    0.8208  &   0.5218  &   0.9190\\
& U2Fusion  &    0.8204 &    0.5947  &   0.9430\\
& GMC  &    0.8205 &    0.5518  &   0.9033\\ \hline
\end{tabular}
\end{center}
\caption {The objective results of different methods (these comparisons are only possible for single channel image pairs).  Metrics: Wang ($Q_0$ \cite{wang2008performance}), Xydeas ($P_e$ \cite{2000Objective}), Piella ($Q$ \cite{piella2003new}).  FM0-3 defined in section 4.1.  Comparison techniques:  U2Fusion \cite{xu2020u2fusion} and GMC \cite{anantrasirichai2020image}.} 
\label{tab 1} 
\end{table}

\begin{table}[!htb]
\begin{center}
\small
\begin{tabular}{cccc}
\hline
\textbf{Fusion Pair} & \textbf{$c_0$, P($c_0$)} & \textbf{$c_1$, P($c_1$)}  & \textbf{$c_{FM3}$, P($c_{FM3}$)}  \\ \hline
\multirow{2}{*}{Clocks (Top row figure \ref{fig:fusion})}& Analog\_Clock,  & Analog\_Clock,& Analog\_Clock,\\  
& 0.524  &    0.722  &   0.663 \\
\multirow{2}{*}{Beagle/Ferret (Second row figure \ref{fig:fusion})}& Beagle,  & Black-Footed\_Ferret,& Beagle,\\  
& 0.970  &    0.449  &   0.927\\
\multirow{2}{*}{Great\_Pyrenees/Bee-Eater (Third row figure \ref{fig:fusion})}& Great\_Pyrenees,  & Bee-Eater, & Great\_Pyrenees,\\  
& 0.858  &  0.991  &   0.656\\
\multirow{2}{*}{Goldfinch/Indigo\_bunting (Fourth row figure \ref{fig:fusion})}& Goldfinch,  & Indigo\_Bunting, & Goldfinch,\\  
& 0.9999  &    0.999  &   0.780\\ 
\hline
\end{tabular}
\end{center}
\caption {VGG19 top classification results (class and probability) for input and FM3 fused images (from figure \ref{fig:fusion}).} 
\label{tab 2s} 
\end{table}

\begin{figure*}
\centering
\begin{subfigure}[b]{.124\linewidth}
  \centering
  \includegraphics[width=\linewidth]{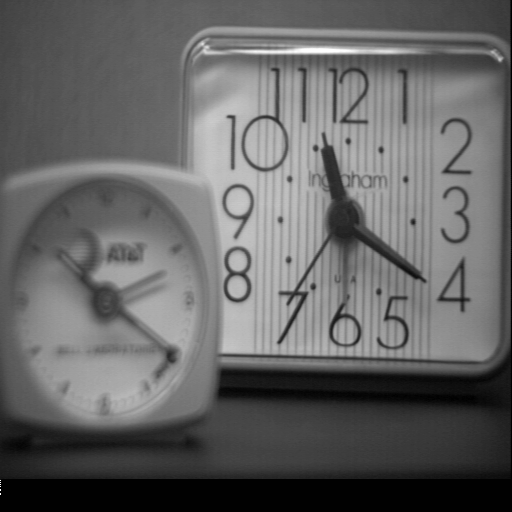}
\end{subfigure}%
\begin{subfigure}[b]{.124\linewidth}
  \centering
  \includegraphics[width=\linewidth]{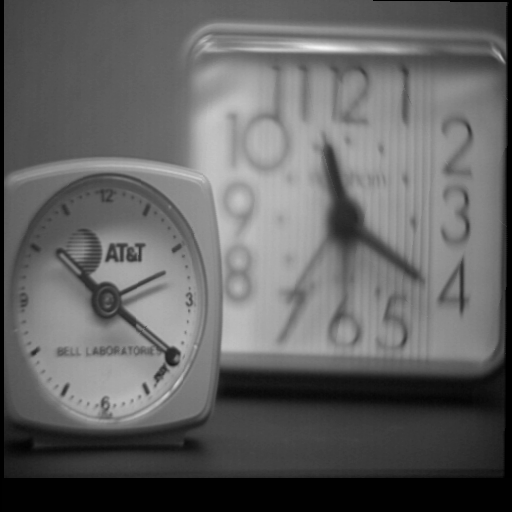}
\end{subfigure}%
\begin{subfigure}[b]{.124\linewidth}
  \centering
  \includegraphics[width=\linewidth]{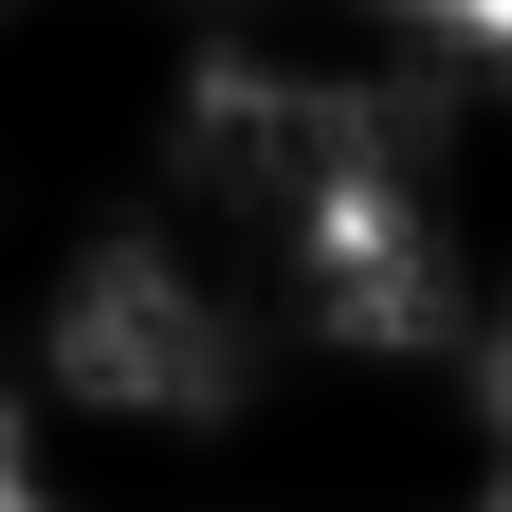}
\end{subfigure}%
\begin{subfigure}[b]{.124\linewidth}
  \centering
  \includegraphics[width=\linewidth]{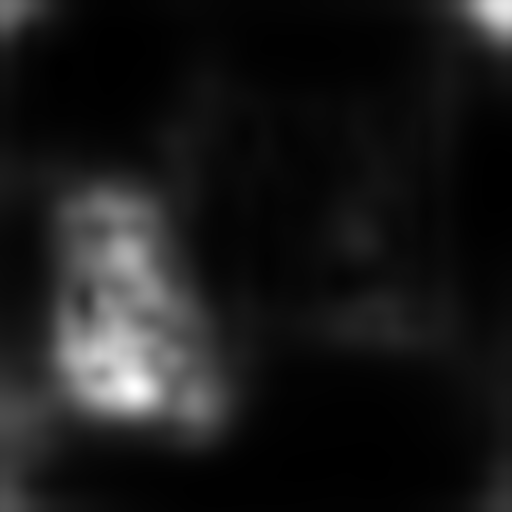}
\end{subfigure}%
\begin{subfigure}[b]{.124\linewidth}
  \centering
  \includegraphics[width=\linewidth]{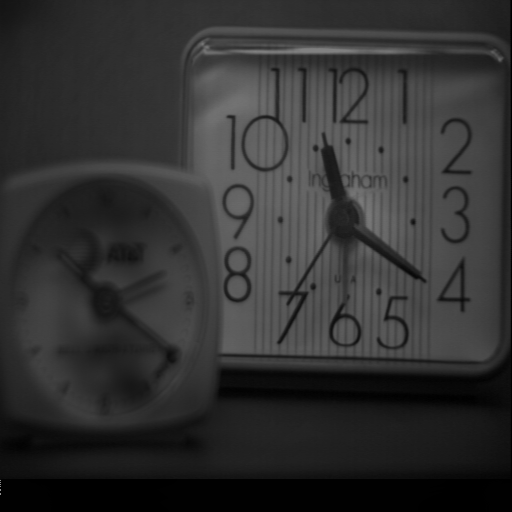}
\end{subfigure}%
\begin{subfigure}[b]{.124\linewidth}
  \centering
  \includegraphics[width=\linewidth]{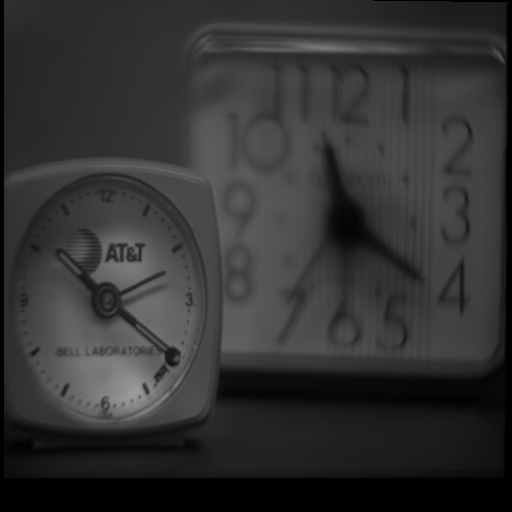}
\end{subfigure}%
\begin{subfigure}[b]{.124\linewidth}
  \centering
  \includegraphics[width=\linewidth]{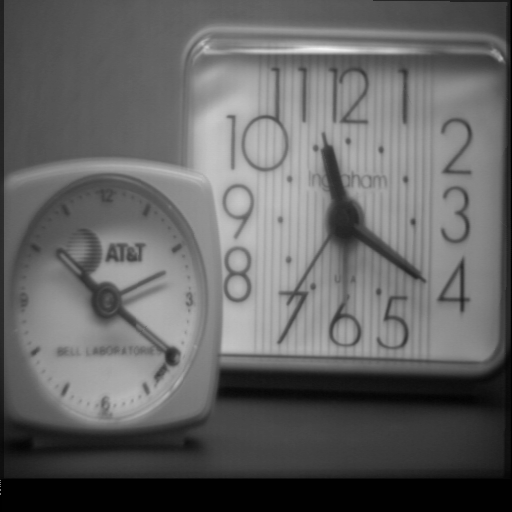}
\end{subfigure}%
\begin{subfigure}[b]{.124\linewidth}
  \centering
  \includegraphics[width=\linewidth]{./OutIms/FIm4/Pair1.png}
\end{subfigure}\\

\begin{subfigure}[b]{.124\linewidth}
  \centering
  \includegraphics[width=\linewidth]{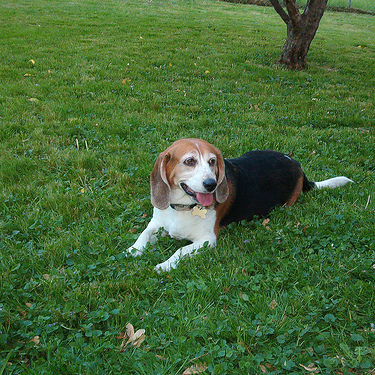}
\end{subfigure}%
\begin{subfigure}[b]{.124\linewidth}
  \centering
  \includegraphics[width=\linewidth]{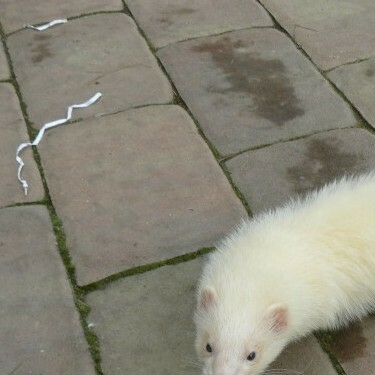}
\end{subfigure}%
\begin{subfigure}[b]{.124\linewidth}
  \centering
  \includegraphics[width=\linewidth]{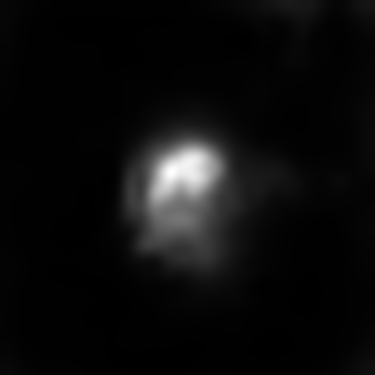}
\end{subfigure}%
\begin{subfigure}[b]{.124\linewidth}
  \centering
  \includegraphics[width=\linewidth]{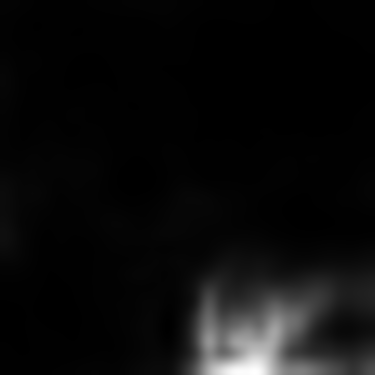}
\end{subfigure}%
\begin{subfigure}[b]{.124\linewidth}
  \centering
  \includegraphics[width=\linewidth]{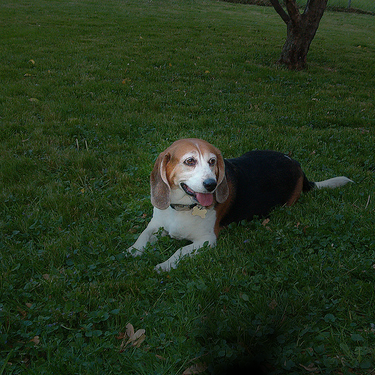}
\end{subfigure}%
\begin{subfigure}[b]{.124\linewidth}
  \centering
  \includegraphics[width=\linewidth]{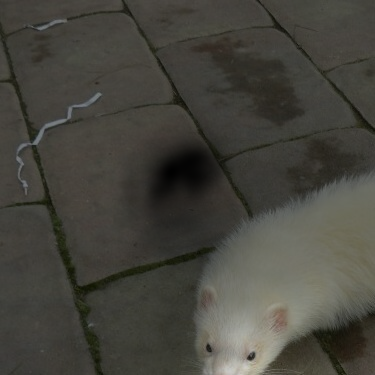}
\end{subfigure}%
\begin{subfigure}[b]{.124\linewidth}
  \centering
  \includegraphics[width=\linewidth]{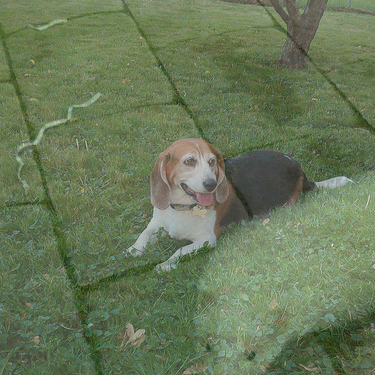}
\end{subfigure}%
\begin{subfigure}[b]{.124\linewidth}
  \centering
  \includegraphics[width=\linewidth]{./OutIms/FIm4/Pair16.png}
\end{subfigure}\\
\begin{subfigure}[b]{.124\linewidth}
  \centering
  \includegraphics[width=\linewidth]{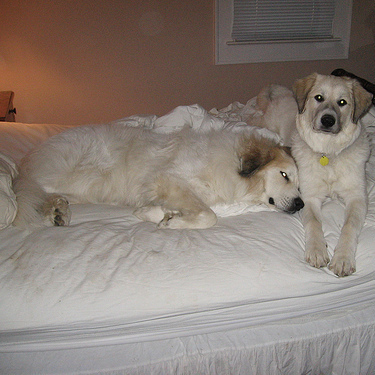}
\end{subfigure}%
\begin{subfigure}[b]{.124\linewidth}
  \centering
  \includegraphics[width=\linewidth]{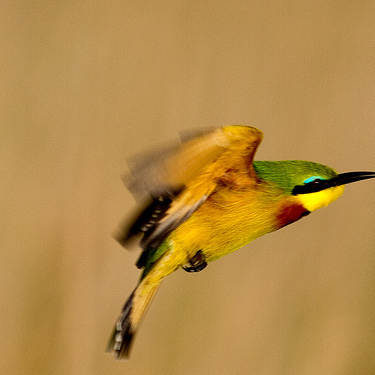}
\end{subfigure}%
\begin{subfigure}[b]{.124\linewidth}
  \centering
  \includegraphics[width=\linewidth]{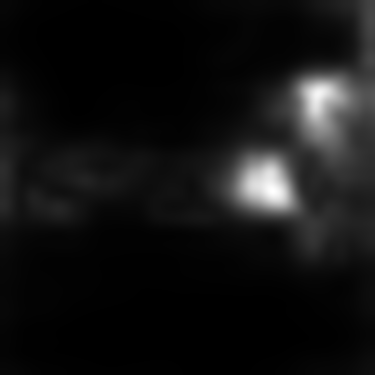}
\end{subfigure}%
\begin{subfigure}[b]{.124\linewidth}
  \centering
  \includegraphics[width=\linewidth]{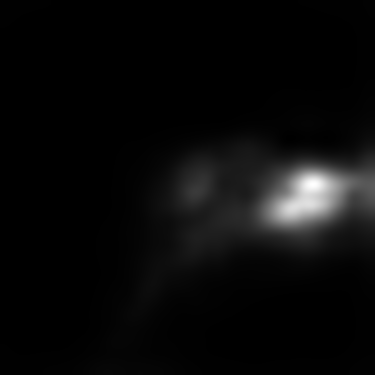}
\end{subfigure}%
\begin{subfigure}[b]{.124\linewidth}
  \centering
  \includegraphics[width=\linewidth]{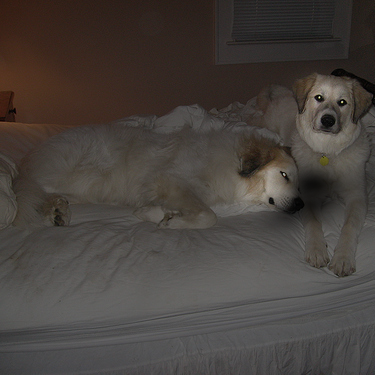}
\end{subfigure}%
\begin{subfigure}[b]{.124\linewidth}
  \centering
  \includegraphics[width=\linewidth]{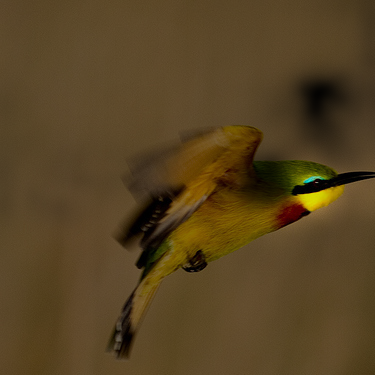}
\end{subfigure}%
\begin{subfigure}[b]{.124\linewidth}
  \centering
  \includegraphics[width=\linewidth]{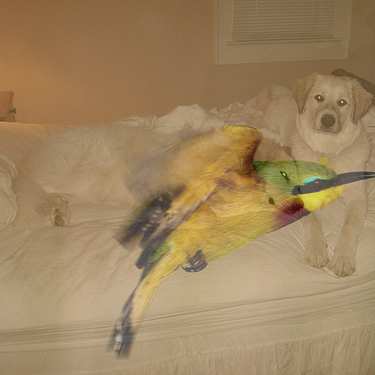}
\end{subfigure}%
\begin{subfigure}[b]{.124\linewidth}
  \centering
  \includegraphics[width=\linewidth]{./OutIms/FIm4/Pair17.png}
\end{subfigure}\\
\begin{subfigure}[b]{.124\linewidth}
  \centering
  \includegraphics[width=\linewidth]{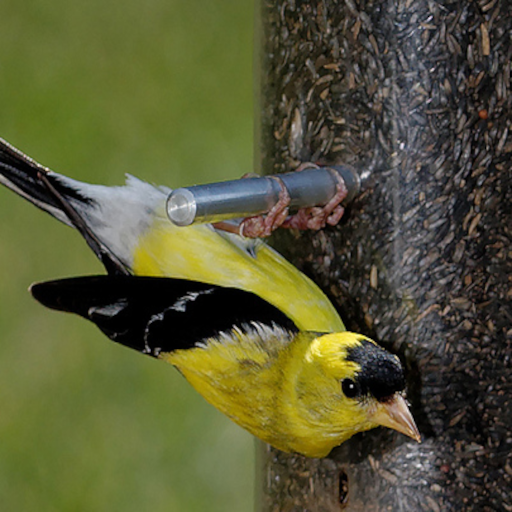}
   \caption*{$I_0$}
\end{subfigure}%
\begin{subfigure}[b]{.124\linewidth}
  \centering
  \includegraphics[width=\linewidth]{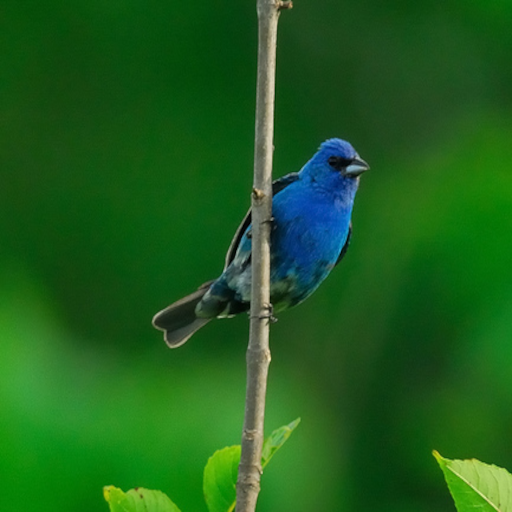}
   \caption*{$I_1$}
\end{subfigure}%
\begin{subfigure}[b]{.124\linewidth}
  \centering
  \includegraphics[width=\linewidth]{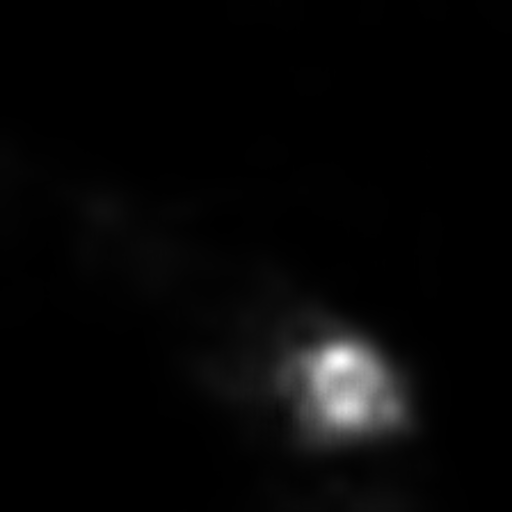}
  \caption*{$P_{c_0,I_0}$}
\end{subfigure}%
\begin{subfigure}[b]{.124\linewidth}
  \centering
  \includegraphics[width=\linewidth]{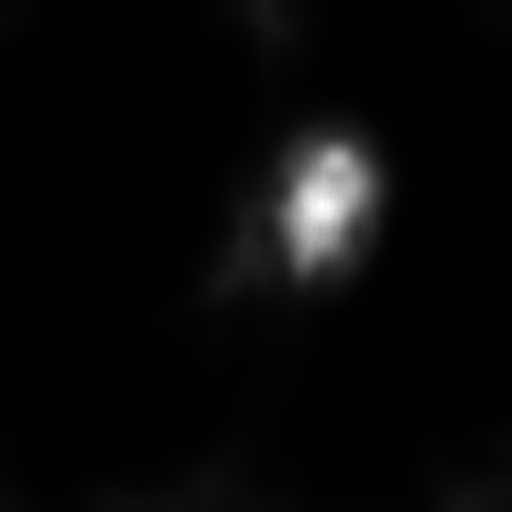}
  \caption*{$P_{c_1,I_1}$}
\end{subfigure}%
\begin{subfigure}[b]{.124\linewidth}
  \centering
  \includegraphics[width=\linewidth]{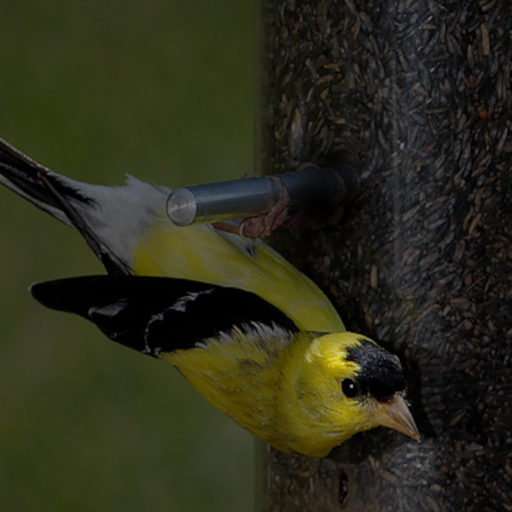}
  \caption*{$P_{M_0}\times I_0$}
\end{subfigure}%
\begin{subfigure}[b]{.124\linewidth}
  \centering
  \includegraphics[width=\linewidth]{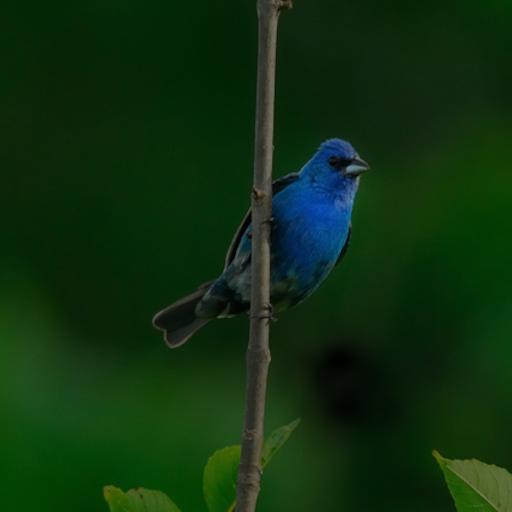}
  \caption*{$P_{M_1}\times I_1$}
\end{subfigure}%
\begin{subfigure}[b]{.124\linewidth}
  \centering
  \includegraphics[width=\linewidth]{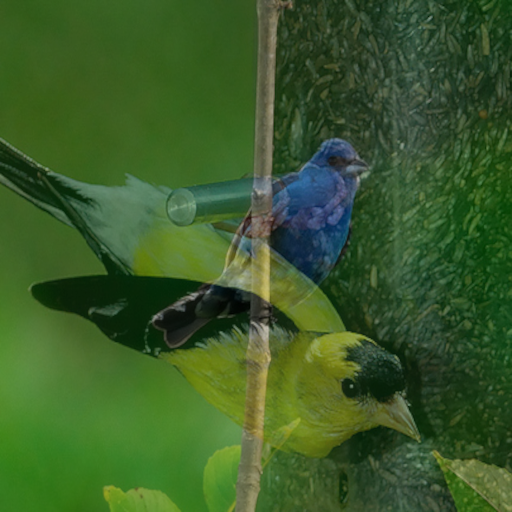}
  \caption*{($(I_0+I_1)/2$)}
\end{subfigure}%
\begin{subfigure}[b]{.124\linewidth}
  \centering
  \includegraphics[width=\linewidth]{./OutIms/FIm4/Pair2.png}
  \caption*{FM3}
\end{subfigure}

\caption{Fused image results.  From top to bottom the fusion pairs are labelled
"Clock", "Beagle/Ferret", "Great\_Pyrenees/Bee-Eater" and "Goldfinch/Indigo\_bunting".  This image shows the input image, CAM mappings, the CAM based mixtures from each image, the averaged image output and the fused image output (FM3).}
\label{fig:fusion}
\end{figure*}
\section{Conclusion}
This paper has defined four novel fusion methods that utilise: image optimisation; choose maximum and majority filter fusion rules; and class activation mappings for semantic fusion.  The image optimisation method was able to achieve state of the art image fusion results measured using conventional image fusion metrics in multiple domains.   This method is also  directly extendable to colour images (not historically a major focus of image fusion). 

Furthermore the CAM based methods can, for the first time, directly exploit semantic information from the top classified class in each of the fused images to generate true semantic level image fusion.  It is conjectured that combining regions that are important to the classification of a set of images will most effectively combine the semantic information from the input images.  Images combined in this semantically meaningful way are shown to retain the important semantic information from both images.

This method would easily extend to multiple images and the fusion of the top-5 classes in each image.
\bibliographystyle{IEEEtranN}
\bibliography{semfuse}
\end{document}